\documentclass[10pt,twocolumn,letterpaper]{article}

\usepackage{cvpr}
\usepackage{times}
\usepackage[utf8]{inputenc}
\usepackage{epsfig}
\usepackage{graphicx}
\usepackage{xcolor,colortbl}
\usepackage{subfig}
\usepackage{amssymb}
\usepackage{amsmath}
\usepackage{pifont}
\usepackage{array}
\usepackage{bbold}
\usepackage{nicefrac}
\usepackage{dsfont}
\usepackage{csquotes}
\usepackage{booktabs}
\usepackage{multirow}
\usepackage{slashbox}
\usepackage{makecell}
\usepackage{siunitx}
\usepackage[font=small]{caption}

\newdimen\owntablesep
\newcommand{\sign}{{\rm sign}}

\newcommand{\putindex}[3]{\vtop{\hbox{\hspace{#3} $#1$}
            \hbox{\raise 6mm \hbox{$\scriptscriptstyle #2$}}}}

\newcommand{\gradx}[0]{\vtop{\hbox{\rm grad}
            \hbox{\raise 2.5mm \hbox{\rm \hspace{2mm} \footnotesize x}}}}

\newcommand{\grady}[0]{\vtop{\hbox{\rm grad}
            \hbox{\raise 2.5mm \hbox{\rm \hspace{2mm} \footnotesize y}}}}

\newcommand{\grad}[1]{\vtop{\hbox{\rm grad}
            \hbox{\raise 2.5mm \hbox{#1}}}}

\newcommand{\btb}{     \begin{tabbing}             }
\newcommand{\bte}{     \end{tabbing}               }

\DeclareUnicodeCharacter{2212}{-}
\DeclareMathOperator*{\argmax}{argmax}

\usepackage[pagebackref=true,breaklinks=true,letterpaper=true,colorlinks,bookmarks=false]{hyperref}
\DeclareSymbolFont{AMSb}{U}{msb}{m}{n}
\DeclareSymbolFontAlphabet{\mathbb}{AMSb}

\cvprfinalcopy 

\def\cvprPaperID{2} 

\ifcvprfinal\fi
\begin{document}

\title{Improved Noise and Attack Robustness for Semantic Segmentation\\ by Using Multi-Task Training with Self-Supervised Depth Estimation}

\author{Marvin Klingner\quad Andreas Bär\quad Tim Fingscheidt\\
{\tt\small \{m.klingner, andreas.baer, t.fingscheidt\}@tu-bs.de}\\[1.0em]
Technische Universität Braunschweig}

\maketitle
\thispagestyle{empty}

\begin{abstract}
While current approaches for neural network training often aim at improving performance, less focus is put on training methods aiming at robustness towards varying noise conditions or directed attacks by adversarial examples. In this paper, we propose to improve robustness by a multi-task training, which extends supervised semantic segmentation by a self-supervised monocular depth estimation on unlabeled videos. This additional task is only performed during training to improve the semantic segmentation model's robustness at test time under several input perturbations. Moreover, we even find that our joint training approach also improves the performance of the model on the original (supervised) semantic segmentation task. Our evaluation exhibits a particular novelty in that it allows to mutually compare the effect of input noises and adversarial attacks on the robustness of the semantic segmentation. We show the effectiveness of our method on the Cityscapes dataset, where our multi-task training approach consistently outperforms the single-task semantic segmentation baseline in terms of both robustness vs.\ noise and in terms of adversarial attacks, without the need for depth labels in training.
\end{abstract}

\section{Introduction}

The task of semantic segmentation is one of the major challenges for robust environment perception in autonomous driving. While most current approaches aim at improving the performance of such models \cite{Chen2018, Chen2018a, Orsic2019, Zhu2019}, little focus has been put on training models that are robust to input perturbations. These perturbations can occur as noise, or they may stem from using different cameras, or even they can be targeted attacks \cite{Metzen2017, Moosavi-Dezfooli2017} from someone aiming at corrupting the behavior of the semantic segmentation. Particularly such directed attacks are a major safety issue when deploying deep neural networks (DNNs) and they have been introduced in various forms ranging from white-box single-image attacks \cite{Carlini2017a, Goodfellow2015, Madry2018} to black-box universally applicable attacks \cite{Moosavi-Dezfooli2017, Mopuri2017, Liu2019e}. As of now there is no single approach for a model to robustly handle such various input perturbations. Current strategies either rely on augmenting the training material such that the model is trained on the perturbations it should be robust to \cite{Goodfellow2015, Madry2018}, or they aim at implementing a non-differentiable preprocessing on the input such that the perturbations are erased \cite{Guo2018,Liu2019c}, however, resulting in decreasing model performance on unperturbed images. Conclusively, there is still a need for strategies to robustly train neural networks, which we address by incorporating self-supervised depth estimation during training.
\par
\begin{figure}
\centering
\includegraphics[width=1.0\linewidth]{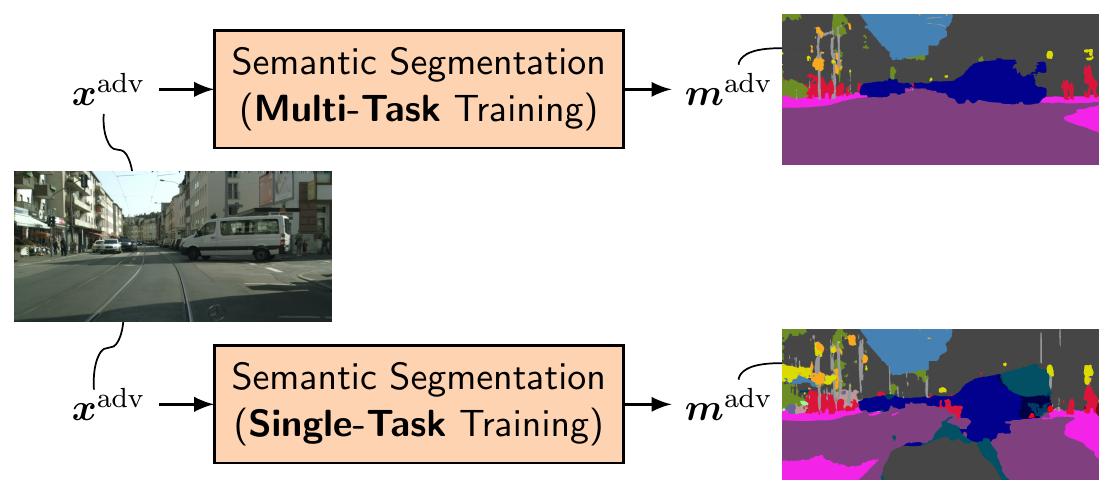}
\caption{Demonstration of the \textbf{effect} of our self-supervised depth learning approach \textbf{on the robustness of semantic segmentation} networks. For adversarial attacks of equal strength the output of a network trained in a multi-task fashion shows superior results while having the same computational complexity at test time. Our multi-task training does not require depth labels.}
\label{fig:main_results}
\end{figure}
Another interesting field is multi-task learning, where several tasks are combined to take profit from the additional knowledge of the respective other task. Several approaches show that combining semantic segmentation with depth estimation \cite{Kendall2018, Zhang2019a} improve the model's performance. These techniques are especially promising for a supervised task, when the other task can be trained in a self-supervised fashion, as thereby the model can be implicitly trained on a very large amount of data. Recent works have combined semantic segmentation with self-supervised depth estimation from stereo images \cite{Chen2019a, Novosel2019}, where particularly depth estimation takes profit from the semantic scene representation knowledge. In this work, we utilize self-supervised monocular depth estimation from unlabeled sequential images (videos), eliminating the need for stereo image pairs, to improve the robustness of our semantic segmentation model --- without the need of depth labels.\par
While it is a well-known fact that multi-task learning can improve the performance of semantic segmentation models \cite{Kendall2018, Vu2019a, Yang2018b}, it is yet unclear, whether also robustness can be improved by such techniques. A recent approach from classification \cite{Hendrycks2019a} suggests that additional training of self-supervised tasks can indeed improve the robustness of such models, while there is no such indication for pixel-level tasks until now. In this work we show the positive effect of multi-task learning with self-supervised depth estimation on semantic segmentation models in terms of robustness. Also our work suggests that this beneficial effect can even be observed when both tasks are trained on different datasets, making our technique widely applicable. \par
To sum up, our contributions are the following: Firstly, we improve the performance of a semantic segmentation model by multi-task learning on unlabeled videos eliminating the need for stereo image pairs. Secondly, we analyze our multi-task learning framework not only in terms of performance but also in terms of robustness to input perturbations. Thirdly, we thereby develop an easy-to-generalize self-supervised strategy to increase a model's robustness to input perturbations.\par
The paper is structured as follows: First we give an overview over related work in Section \ref{sec:2}, followed by our method description in Section \ref{sec:3}. In Section \ref{sec:4} we describe our experimental setup, which we evaluate in Section \ref{sec:5}, finally concluding the paper in Section \ref{sec:6}.

\section{Related Work}
\label{sec:2}
In this section we start by giving an overview over relevant approaches in semantic segmentation and self-supervised depth estimation and how the latter task is beneficial in multi-task learning setups. Finally, we will discuss other works on robustness of neural networks.
\par
\textbf{Semantic Segmentation}: 
Semantic segmentation aims at the pixel-wise classification of an image. Early work done by Long \etal~\cite{Long2015} unveiled the superiority of fully-convolutional neural networks (FCNs) for this task. Building upon the concept of FCNs, subsequent work concentrated on the additional use of dilated convolutions \cite{Chen2015, Yu2016}, recurrent neural networks (RNNs) in spatial direction \cite{Zhuang2018}, unsupervised domain adaptation techniques \cite{Bolte2019a, Zou2018}, forms of spatial pyramid pooling \cite{Chen2018, Zhao2016a}, different schemes for intermediate skip-connections \cite{Bilinski2018, Chen2018a}, state-of-the-art feature extractors as backbones \cite{Chen2018, Orsic2019}, post-processing with conditional random fields (CRFs) \cite{Chen2018, Vemulapalli2016} and label relaxation \cite{Zhu2019}, and multi-scale inference \cite{Chen2018, Zhao2016a}, for further performance improvements.
\par
Recently, the focus of research switched from a performance point of view to an efficiency point of view.
Efficiency can be achieved by factorizing the convolution operation \cite{Mehta2018, Romera2018}, combining depthwise-separable convolution \cite{Chollet2017} with inverted residual units \cite{Sandler2018}, using in-place operations \cite{RotaBulo2018}, or a more efficient architectural design \cite{Li2019a, Orsic2019}.
\par
\textbf{Self-Supervised Depth Estimation}:
Depth estimation describes the task of assigning each pixel its distance from the camera, which has been successfully approached by Eigen \etal~\cite{Eigen2014} by training a neural network on labels generated by a LiDAR sensor, with consecutive works improving the technique \cite{Fu2018, Guizilini2019a}. More recently, the problem has also been reformulated to a self-supervised setting \cite{Garg2016}, where the depth is optimized as the parameter of an image reprojection model. While initial approaches relied on stereo image pairs \cite{Garg2016, Godard2017}, Zhou \etal~\cite{Zhou2017a} showed the applicability of the technique to image sequences. This generally applicable technique has been drastically improved by better reprojection losses \cite{Casser2019, Godard2019, Mahjourian2018}, increasing the image resolution \cite{Pillai2019} or utilizing synthetic data \cite{Bozorgtabar2019}. The work of Gordon \etal~\cite{Gordon2019} does not even rely on the knowledge of any camera parameters. Current state-of-the-art algorithms for self-supervised monocular depth estimation \cite{Godard2019, Guizilini2019} are closing the gap to fully-supervised methods and can easily make use of almost unlimited amounts of data. In this work, our network architecture as well as our training approach for self-supervised depth estimation training approach is based on \cite{Godard2019}, a state-of-the-art framework for self-supervised monocular depth estimation.
\par
\textbf{Multi-Task Learning}: 
Recent work shows that training depth estimation jointly with other tasks such as optical flow \cite{Chen2019b, Zou2018a}, semantic segmentation \cite{Kendall2018, Xu2018c, Zhang2019a}, or domain adaptation \cite{Vu2019a} improves the other task's performance due to a more generalized learned scene understanding. Particularly the incorporation of semantic segmentation into the self-supervised depth estimation has been utilized to identify moving objects \cite{Casser2019, Meng2019a}, whose depth estimation could thereby be improved. Other works utilize the semantic segmentation in a multi-task learning setting mainly aiming at improving the self-supervised depth estimation \cite{Guizilini2020, Yang2018b}. In our work the focus is on the semantic segmentation task rather than on the depth estimation task. We also focus on robustness rather than on absolute performance.\par
There have been also initial works on combining self-supervised depth estimation from stereo image pairs with semantic segmentation. The work of \cite{Chen2019a} shows that in this case the semantic segmentation takes profit from multi-task training with self-supervised depth estimation, while recently Novosel \etal could show the same effect \cite{Novosel2019}. Our approach transfers this setting to more general self-supervised learning from monocular image sequences (eliminating the need for stereo image pairs), since the main focus of this work is the improved robustness of such models trained in a multi-task fashion.
\par
\textbf{Robustness}:
The influence of additive noise in training DNNs is well known \cite{Bishop1995,Holstroem1992}. Recent work showed the existence of adversarial examples \cite{Szegedy2014}, a special kind of visually imperceptible image perturbation leading to complete deception of an underlying DNN. This observation led to subsequent work trying to efficiently craft image-specific targeted or non-targeted adversarial examples iteratively or non-iteratively, e.g., the fast gradient sign method (FGSM) \cite{Goodfellow2015}, the Carlini and Wagner attack (CW) \cite{Carlini2017a}, the momentum iterative fast gradient sign method (MI-FGSM) \cite{Dong2018a}, and projected gradient descent (PGD) \cite{Madry2018}. Besides that, other works aim at finding image-agnostic adversarial perturbations, e.g., universal adversarial perturbation (UAP) \cite{Moosavi-Dezfooli2017}, fast feature fool (FFF) \cite{Mopuri2017}, and prior-driven uncertainty estimation (PD-UA) \cite{Liu2019e}. While most works were initially proposed for image classification, their application to more complex tasks such as semantic segmentation is possible \cite{Arnab2018, Mopuri2018}. Specifically designed attacks for semantic segmentation lead to even more realistic outcomes in terms of spatial consistency of the segmentation mask \cite{Assion2019, Metzen2017}.
\par
DNNs' lack of robustness towards adversarial attacks motivated research in defense methods. They mostly build upon adversarial training \cite{Goodfellow2015, Madry2018}, robustness-oriented loss functions \cite{Baer2019, Chen2019e}, self-supervision \cite{Hendrycks2019a}, new intermediate layers to tackle feature noise \cite{He2019c}, or different kinds of pre-processing strategies \cite{Guo2018, Liu2019c}. Nonetheless, the latter methods mostly rely on gradient masking which can be circumvented \cite{Athalye2018}, proving that there is still a need for strategies to improve robustness of DNNs, as addressed in this work.

\section{Theoretical Background}
\label{sec:3}
Here, we introduce our framework for the joint training of semantic segmentation and self-supervised depth estimation. Afterwards, we describe our inference setting, where we apply a perturbation on an input image and measure how robust the model is with respect to this perturbation.

\begin{figure}[t]
	\centering	
	\includegraphics[width=0.94\linewidth]{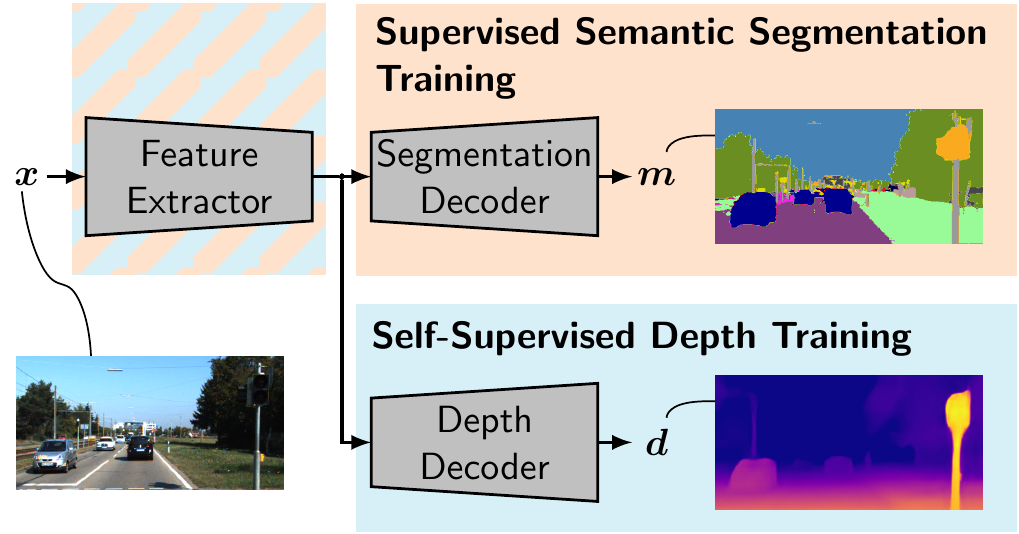}
	\caption{\textbf{Our multi-task training setup}, where we use a second decoder head for multi-task learning with self-supervised depth estimation.}
	\label{fig:overview_training}
\end{figure} 

\subsection{Semantic Segmentation}

The input of our semantic segmentation network is an image $\boldsymbol{x} \in \mathcal{G}^{H\times W\times C}$ (cf.\ Fig.\ \ref{fig:overview_training}), where $\mathcal{G} = \left\lbrace 0, 1, ..., 255 \right\rbrace$ represents the set of gray values and $H$, $W$, and $C=3$ represent the height, the width and the number of color channels of an image, respectively. As semantic segmentation is the task of assigning a class label to each image color pixel $\boldsymbol{x}_i\in \mathcal{G}^C$, with $i \in \mathcal{I} = \left\lbrace 1, ..., H\cdot W\right\rbrace$, the input image is converted to output probabilities $\boldsymbol{y} \in \mathbb{I}^{H\times W\times |\mathcal{S}|}$ by a neural network, where $\mathbb{I} = \left[0, 1\right]$ and $|\mathcal{S}|$ is the number of semantic classes the model can predict. Each element $y_{i,s}$ of $\boldsymbol{y}$ can be interpreted as the posterior probability that an image pixel $\boldsymbol{x}_i$ belongs to class $\mathcal{S} =  \left\lbrace1,2,...,|\mathcal{S}|\right\rbrace$. The final segmentation mask $\boldsymbol{m}\in\mathcal{S}^{H\times W}$ can be computed by assigning a class label to each image pixel using $m_{i} = \argmax_{s \in \mathcal{S}} y_{i, s}$. The network is optimized by the weighted cross-entropy loss function
\begin{equation}
J^{\mathrm{ce}} = -\frac{1}{|\mathcal{I}|}\sum_{i \in\mathcal{I}}\sum_{s \in\mathcal{S}} w_s \overline{y}_{i,s} \cdot \log\left(y_{i,s}\right), 
\label{eq:crossentropy_loss}
\end{equation}
where $w_s$ are class-wise computed weights as in \cite{Paszke2016}, and $\overline{y}_{i,s}$ are the elements of the one-hot encoded class labels $\overline{\boldsymbol{y}}$, and $|\mathcal{I}|=HW$ is the number of pixels. Notice that the ground truth segmentation mask $\overline{\boldsymbol{m}}\in \mathcal{S}^{H\times W}$ can be computed in a straightforward manner element-wise by $\overline{m}_{i} = \argmax_{s \in \mathcal{S}} \overline{y}_{i, s}$.

\subsection{Self-Supervised Depth Estimation}

To increase the amount of data the network has been trained on, we extend our training setup by a second decoder head providing pixel-wise depth estimates $d_i \in \mathbb{D}$ as shown in Figure \ref{fig:overview_training}. The range $\mathbb{D} = \left[d_{\mathrm{min}},d_{\mathrm{max}}\right]$ is defined by a lower bound $d_{\mathrm{min}}$ and an upper bound $d_{\mathrm{max}}$ of possible depth values. Conclusively, the network's output is a dense depth map $\boldsymbol{d}\in \mathbb{D}^{H\times W}$.\par
Optimizing the network requires consecutive pairs of images $\boldsymbol{x}_t$, $\boldsymbol{x}_{t'}$ with $t'\in\mathcal{T}' = \left\lbrace {t\!-\!1}, {t\!+\!1} \right\rbrace$, which are fed to a pose network, predicting the two relative poses $\boldsymbol{T}_{t\rightarrow t'}\in \mathit{SE}(3)$ between $\boldsymbol{x}_t$ and $\boldsymbol{x}_{t'}$, $t' \in \mathcal{T}'$, with $\mathit{SE}(3)$ being the special Euclidean group representing all possible rotations and translations \cite{Szeliski2010}. Additionally, the depth $\boldsymbol{d}_t$ with respect to the input image $\boldsymbol{x}_t$ is predicted. The predictions $\boldsymbol{T}_{t\rightarrow t'}$ and $\boldsymbol{d}_t$ can be used to project the image frames $\boldsymbol{x}_{t'}$ at time $t'$ to the pixel coordinates of the image $\boldsymbol{x}_t$ \cite{Chen2019b} yielding the two projected images $\boldsymbol{x}_{t'\rightarrow t}$, $t'\in\mathcal{T}'$. Note that as the projected images depend on the network predictions $\boldsymbol{d}_t$ and $\boldsymbol{T}_{t\rightarrow t'}$, the optimization can be done by optimizing the pixel-wise distance between the current frame's color pixels $\boldsymbol{x}_{t,i}$ and the color pixels of the projected frame $\boldsymbol{x}_{t'\rightarrow t, i}$, implicitly optimizing $\boldsymbol{d}_t$ and $\boldsymbol{T}_{t\rightarrow t'}$ by optimizing the image projection model. Adopting the approaches of \cite{Casser2019, Yin2018}, we calculate a weighted sum of the absolute difference and the structural similarity (SSIM) difference \cite{wang2004image} with a weighting factor $\alpha=0.85$. To robustly handle occlusions between consecutive frames, we also adopt the photometric (ph) \textit{per-pixel minimum} reprojection loss of \cite{Godard2019}, yielding
\begin{align}
	J_t^{\mathrm{ph}} &=  \frac{1}{|\mathcal{I}|}\sum_{i \in\mathcal{I}} \min_{t'\in\mathcal{T}'} \left( \frac{\alpha}{2}\left(1-\mathrm{SSIM}_i\left(\boldsymbol{x}_{t}, \boldsymbol{x}_{t'\rightarrow t}\right)\right)\right. \notag\\ &+  \left(1-\alpha\right) \frac{1}{C}\left\lVert\boldsymbol{x}_{t,i} - \boldsymbol{x}_{t'\rightarrow t, i}\right\rVert_1 \Big).
	\label{eq:photometric_loss} 
\end{align}
Here, the SSIM difference at pixel index $i$ is $\mathrm{SSIM}_i\left(\cdot\right) \in \mathbb{I}$, with $\mathbb{I} = \left[0, 1\right]$, and is computed on $3\times 3$ patches of the images $\boldsymbol{x}_{t}$ and $\boldsymbol{x}_{t'\rightarrow t}$. The $L_1$ norm $\left\lVert\cdot \right\rVert_1$ is computed over all $C=3$ gray value channels.\par
To enforce the depth $\boldsymbol{d}_t$ to be only non-smooth in image regions, where there is also a strong gradient inside the gray values, an additional smoothness loss $J_t^{\mathrm{sm}}$ \cite{Godard2017} is applied on the mean-normalized inverse depth $\overline{\boldsymbol{\rho}}_t\in \mathbb{R}^{H\times W}$. Its elements are defined by $\overline{\rho}_{t,i} = \frac{\rho_{t,i}}{\frac{1}{HW} \sum_{j\in\mathcal{I}} \rho_{t,j}}$ with $\rho_{t,i} = \frac{1}{d_{t,i}}$. Conclusively, we get
\begin{align}
	J_t^{\mathrm{sm}} &= \frac{1}{|\mathcal{I}|}\sum_{i \in\mathcal{I}}\left(|\partial_h \overline{\boldsymbol{\rho}}_{t,i}|\exp\left(-\frac{1}{C}\left\lVert\partial_h \boldsymbol{x}_{t,i}\right\rVert_1\right)\right.\notag\\ &+ \left.|\partial_w \overline{\boldsymbol{\rho}}_{t,i}|\exp\left(-\frac{1}{C}\left\lVert\partial_w \boldsymbol{x}_{t,i}\right\rVert_1\right)\right),
	\label{eq:smoothness_loss}
\end{align}
where $\partial_h$ and $\partial_w$ denote the one-dimensional difference quotient with respect to the height and width dimension of the image, for details see \cite{Godard2017}. The complete depth loss can be defined by 
\begin{equation}
	J_t^{\mathrm{depth}} = J_t^{\mathrm{ph}} + \beta J_t^{\mathrm{sm}},
	\label{eq:depth_loss}
\end{equation}
with $\beta = 10^{-3}$ set according to previous works \cite{Casser2019, Godard2019}. 

\subsection{Multi-Task Training}

To train our model end-to-end, we chose a setup with a single shared encoder and two decoder heads, see Figure \ref{fig:overview_training}, each being either trained on depth estimation according to (\ref{eq:depth_loss}), or on semantic segmentation according to (\ref{eq:crossentropy_loss}). During optimization, instead of weighing the two loss functions (\ref{eq:depth_loss}) and (\ref{eq:crossentropy_loss}) against each other, we followed \cite{Ganin2015} in letting the gradients propagate unscaled inside their respective decoder heads and scale them when they reach the shared encoder. Thereby, the decoder heads are optimized independently, while we have full control on how much the gradients of each task should contribute inside the shared encoder. This way, while the encoder learns to extract features for the semantic segmentation task, it also takes profit from the variety of the unlabeled videos used during training of the depth estimation. Assuming gradients $\boldsymbol{g}^{\mathrm{depth}}$ and $\boldsymbol{g}^{\mathrm{seg}}$, which are computed inside the task-specific decoders, the total gradient for further backpropagation into the encoder (i.e., the feature extractor) is computed as
\begin{equation}
	\boldsymbol{g}^{\mathrm{total}} = \left(1-\lambda\right) \boldsymbol{g}^{\mathrm{depth}} + \lambda \boldsymbol{g}^{\mathrm{seg}},
	\label{eq:gradient_scaling}
\end{equation}
with $\lambda$ determining the inter-task gradient scaling factor.

\begin{figure}[t]
	\centering	
	\includegraphics[width=0.94\linewidth]{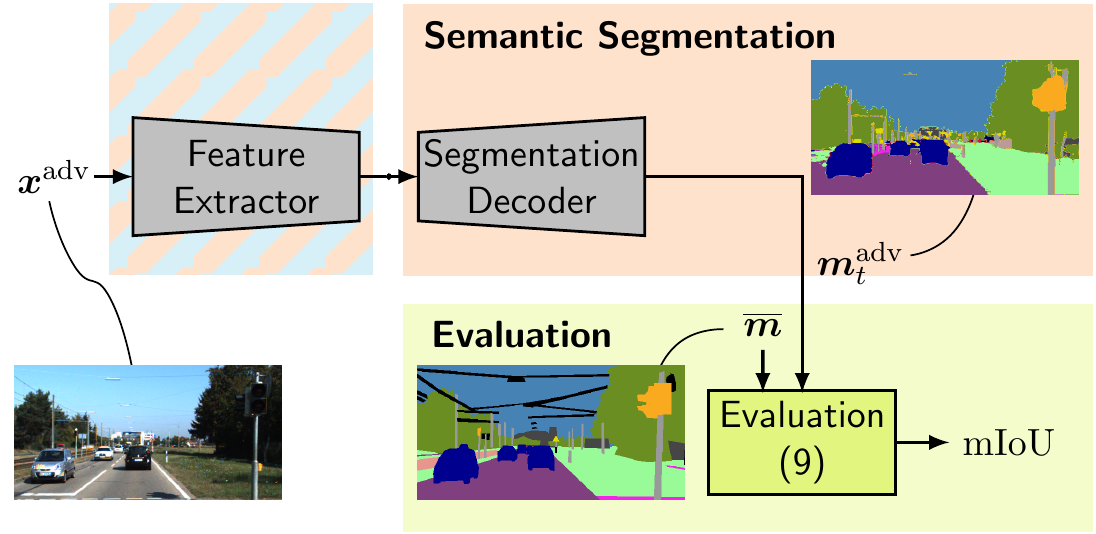}
	\caption{\textbf{Our inference setup}, where the depth decoder head is discarded and only the semantic segmentation is evaluated on input images $\boldsymbol{x}^{\mathrm{adv}}$ that have been perturbed.}
	\label{fig:overview_inference}
\end{figure}

\subsection{Input Perturbation During Inference}

While we optimize both depth estimation and semantic segmentation jointly during training, during inference and evaluation we only examine the performance of the semantic segmentation model with respect to several input perturbations as shown in Figure \ref{fig:overview_inference}. In contrast to the training procedure, where the network is trained with unperturbed images $\boldsymbol{x}$, during inference we alter the input with a perturbation $\boldsymbol{r}$ yielding a perturbed input image $\boldsymbol{x}^{\mathrm{adv}}$. We can usually calculate this perturbed image as
\begin{equation}
	\boldsymbol{x}^{\mathrm{adv}} = \boldsymbol{x} + \boldsymbol{r}\!\left(\epsilon\right),
	\label{eq:add_input_perturbation}
\end{equation}
where $\epsilon$ defines the strength of the input perturbation $\boldsymbol{r}\!\left(\epsilon\right)$. In this work we examine Gaussian noise, salt and pepper noise, and adversarial attacks as input perturbations. For these, we ensure that equal perturbation strengths $\epsilon$ of two perturbation types also result in equal signal-to-noise ratios ($\operatorname{SNR}$) on the input image $\boldsymbol{x}^{\mathrm{adv}}$, thereby making different noise and attack types comparable. The signal-to-noise ratio is defined by $\operatorname{SNR} = \frac{\raisebox{-0.35ex}{\text{$\operatorname{E}$}}\left(\left\lVert\boldsymbol{x}\right\rVert^2_2\right)}{\raisebox{-0.35ex}{\text{$\operatorname{E}$}}\left(\left\lVert\boldsymbol{r} \right\rVert^2_2\right)}$ with $\operatorname{E}\left(\left\lVert\boldsymbol{x}\right\rVert^2_2\right)$ being the expectation value of the image's squared gray values of all $C=3$ color channels, and $\operatorname{E}\left(\left\lVert\boldsymbol{r} \right\rVert^2_2\right)$ being the expectation value of the sum of the squared noise pixels. If we define 
\begin{equation}
\epsilon = \sqrt{\frac{1}{HWC}\operatorname{E}\left(\left\lVert\boldsymbol{r} \right\rVert^2_2\right)},
\label{eq:epsilon}
\end{equation}\par
then for Gaussian noise $\boldsymbol{r}$ sampled from a normal distribution $\boldsymbol{r} \sim \mathcal{N}^{H\times W\times C}$ with  $\mathcal{N} = \mathcal{N}\left(0,\epsilon^2\right)$ and afterwards added to the input image as described by (\ref{eq:add_input_perturbation}), we obtain the SNR as given above. As the elements of $\boldsymbol{r}$ have an expectation value of $0$, the term $\operatorname{E}\left(\left\lVert\boldsymbol{r} \right\rVert^2_2\right)$ becomes the variance of the noise and therefore in this special case, $\epsilon$ can be identified as the standard deviation of the Gaussian distribution.\par
For salt and pepper noise $\boldsymbol{r}$, a fraction $f$ of pixels is randomly set to $0$ or $255$ with equal probability. Therefore, this perturbation type is the only one, where $\epsilon$ cannot be imposed in advance but has to be calculated from the perturbed and unperturbed images $\boldsymbol{x}^{\mathrm{adv}}$ and $\boldsymbol{x}$, respectively. Therefore, we first calculate the noise by rearranging (\ref{eq:add_input_perturbation}) to yield the input perturbation $\boldsymbol{r}\!\left(\epsilon\right)$. Afterwards, we apply (\ref{eq:epsilon}), yielding the perturbation strength $\epsilon$.\par
While these two perturbation types represent random changes of the camera noise or of the environment, it is also possible that the neural network is willingly attacked. In this work, we first consider the white-box fast gradient sign method (FGSM) attack presented by \cite{Goodfellow2015}, which is defined as a perturbation on the input image as
\begin{equation}
	\boldsymbol{x}^{\mathrm{adv}} = \boldsymbol{x} + \epsilon \cdot \sign\left(\nabla_{\boldsymbol{x}} J^{\mathrm{ce}}\left(\overline{\boldsymbol{y}}, \boldsymbol{y}\left(\boldsymbol{x}\right)\right)\right),
	\label{eq:add_fgsm}
\end{equation}
with $\mathrm{sign}(\cdot)$ being the element-wise signum function and $\nabla_{\boldsymbol{x}}$ being the derivative of the loss function (\ref{eq:crossentropy_loss}) with respect to the unperturbed input image $\boldsymbol{x}$. As here the noise $\boldsymbol{r}$ can only take on values of $r_j = \pm \epsilon$, $j \in \left\lbrace 1, ..., H\cdot W\cdot C\right\rbrace$, again the noise variance is equal to $\epsilon^2$ according to (\ref{eq:epsilon}).\par
As one could argue that FGSM is a rather simple one-step attack, we also consider the stronger PGD attack \cite{Kurakin2017}, which is essentially an iterative version of (\ref{eq:add_fgsm}) and thereby better optimized to fooling a neural network.

\subsection{Evaluation Metrics}

A perturbed input image $\boldsymbol{x}^{\mathrm{adv}}$ processed by the trained neural network will produce a different segmentation mask $\boldsymbol{m}^{\mathrm{adv}}$, whose quality will be typically below the unperturbed segmentation mask $\boldsymbol{m}^{\mathrm{clean}}$. Note that for $\epsilon=0$ we have $\boldsymbol{m}^{\mathrm{adv}} = \boldsymbol{m}^{\mathrm{clean}}$.\par
To evaluate our absolute semantic segmentation performance we utilize the intersection over union metric \cite{Everingham2015}, which is defined by 
\begin{equation}
	\mathrm{mIoU} = \frac{1}{|\mathcal{S}|}\sum_{s\in\mathcal{S}}\frac{\mathrm{TP}_s}{\mathrm{TP}_s + \mathrm{FP}_s + \mathrm{FN}_s},
\end{equation}
with true positive ($\mathrm{TP}_s$), false negatives ($\mathrm{FN}_s$), and false positive ($\mathrm{FP}_s$) predictions for each class $s$, all calculated between the ground truth segmentation mask $\overline{\boldsymbol{m}}$ and the predicted one $\boldsymbol{m}$. Note that $\mathrm{TP}_s$, $\mathrm{FN}_s$ and $\mathrm{FP}_s$ are first summed up over all test images and only afterwards the $\mathrm{mIoU}$ is calculated.
\par
When comparing different models with respect to their robustness towards input perturbations, it is not reasonable to compare them by their absolute performance, as their initial performance might already differ. It is then better to utilize the mIoU ratio $Q$ defined by 

\begin{equation}
	Q = \frac{\mathrm{mIoU}^{\mathrm{adv}}}{\mathrm{mIoU}^{\mathrm{clean}}}, 
\end{equation}

where $\mathrm{mIoU}^{\mathrm{clean}}$ and $\mathrm{mIoU}^{\mathrm{adv}}$ are the mIoU values calculated for segmentation masks of clean images $\boldsymbol{m}^{\mathrm{clean}}$ and the ones for perturbed images $\boldsymbol{m}^{\mathrm{adv}}$, respectively. 

\section{Experimental Setup}
\label{sec:4}
Explaining our experimental setup, which is implemented in \texttt{PyTorch} \cite{Paszke2019}, we start with the employed network architecture, as well as our training procedure. Afterwards, we give an overview over our utilized datasets.\\
\begin{table}[t]
  \small
  \centering
  \setlength{\tabcolsep}{4pt}
  \begin{tabular}{|l|rcc|}
  \hline
  \backslashbox{Dataset}{Subset} & training & validation & testing\\
  \hline
  KITTI & 28,937 & 1,158 & \textbf{200}\\
  Cityscapes & 2,975 & \textbf{500} & 1525\\
   \hline
  \end{tabular}
  \caption{Used subsets of the Cityscapes and KITTI datasets with their exact number of images. The subsets on which we report our results are marked in \textbf{boldface}.}
  \label{tab:comparison_dataset}
\end{table}
\textbf{Network Architecture}:
Our basic encoder-decoder network architecture is adapted from \cite{Godard2019}, where the encoder is a \texttt{ResNet18} model \cite{He2016} which has been pretrained on Imagenet \cite{Russakovsky2015}. While we use a unified encoder for both tasks, we utilize seperate decoder heads having the same architecture, except for the last layer. Here the depth head produces a sigmoid output $\boldsymbol{\sigma}\in \mathbb{I}^{H\times W}$, whose pixel-wise elements $\sigma_i$ are converted to depth values by $\frac{1}{a \sigma_{i} + b}$, with $a$ and $b$ constraining the depth values to the range $\left[0.1, 100\right]$. Meanwhile the segmentation head produces output logits in $S= |\mathcal{S}|$ feature maps which undergo a softmax function for the conversion to class probabilities. Our used pose network is also adapted from \cite{Godard2019}.\par
\textbf{Training Details}:
For training of the depth on KITTI we transform the input images to a resolution $640\times 192$, while we downscale the images from the Cityscapes dataset by a factor of $2$ and then randomly crop them to the same resolution. To augment our training material, we use horizontal flipping as well as random brightness ($\pm 0.2$), contrast ($\pm 0.2$), saturation ($\pm 0.2$), and hue ($\pm 0.1$). Finally, we apply a zero-mean normalization on all input images. Note that the photometric loss (\ref{eq:photometric_loss}) is computed on four scales and on images without color augmentations as in \cite{Godard2019}.\par
We apply the gradient scaling of (\ref{eq:gradient_scaling}) at all connections between encoder and decoder during training. For optimization we use the Adam optimizer \cite{Kingma2015} with a learning rate of $10^{-4}$, which is reduced to $10^{-5}$ after 30 epochs and applied for another 10 epochs. To equal the amount of data all models have seen during training, we use a batch size of $12$ for single-task segmentation models, and batch sizes of $6$ on each dataset for our multi-task models.\par
\textbf{Databases}:
We jointly train both tasks, however, each task on a separate dataset. An overview over the number of images inside our used data subsets is given in Table \ref{tab:comparison_dataset}. While we use the Cityscapes dataset \cite{Cordts2016} to supervisedly train the semantic segmentation task, we simultaneously use the KITTI dataset \cite{Geiger2013} to train our depth estimation model. As there is no predefined training, validation, or test subset on this dataset, we adapt the so-called \textit{KITTI split} defined by \cite{Godard2017}, whose test set consists of the 200 training images from the KITTI Stereo 2015 dataset \cite{Menze2015}. As the \textit{KITTI split} has been originally used for a stereo self-supervised depth estimation model, the number of training images deviates slightly from the original definition as we demand a preceding and succeeding frame during training of the depth estimation. We report our semantic segmentation results on the Cityscapes validation set and the KITTI split's test set.

\section{Experimental Evaluation}
\label{sec:5}
\begin{figure}[t]
	\centering	
	\includegraphics[width=1.0\linewidth]{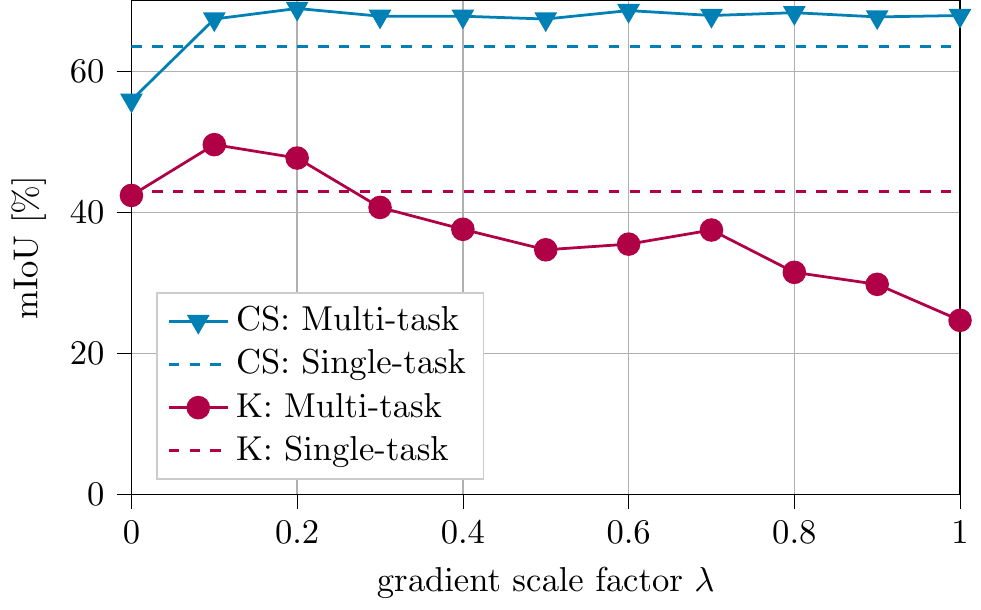}
	\caption{\textbf{Semantic segmentation performance} on the Cityscapes validation set (CS) and on the KITTI split's test set (K) for different scaling factors $\lambda$ in (\ref{eq:gradient_scaling}). Note that we evaluated the same model on two different datasets.}
	\label{fig:scaling_factor}
\end{figure}
\begin{table}[t]
  \small
  \centering
  \setlength{\tabcolsep}{4pt}
  \begin{tabular}{|l|cc|}
  \hline
  Training type & $\mathrm{mIoU}_{\mathrm{CS}}$ & $\mathrm{mIoU}_{\mathrm{K}}$\\
  \hline
  Single-task semantic segmentation & $63.5$ & $43.0$ \\
  Multi-task with depth ($\lambda = 0.1$) & $67.4$ & $\textbf{49.6}$ \\
  Multi-task with depth ($\lambda = 0.2$) & $\textbf{68.9}$ & $47.7$ \\
  Multi-task with depth ($\lambda = 0.5$) & $67.4$ & $34.7$ \\
  Multi-task with depth ($\lambda = 0.9$) & $67.7$ & $29.8$ \\
   \hline
  \end{tabular}
  \caption{\textbf{Semantic segmentation performance} (mIoU values in $\left[\%\right]$) on the Cityscapes validation set (CS) and on the KITTI split's test set (K) with and without multi-task self-supervised depth training for different scale factors $\lambda$. Note that we evaluated the same model on two different datasets.}
  \label{tab:overall_performance}
\end{table}
To show the positive effects of multi-task learning with self-supervised depth estimation from videos, we first show the improved absolute performance in Section \ref{sec:5_1}. Afterwards, we present the main contribution of our work in Sections \ref{sec:5_2} and \ref{sec:5_3}, with the semantic segmentation's robustness to various input perturbations being strongly improved. We provide an ablation study in Section \ref{sec:5_4}. 

\subsection{Evaluation w.r.t. Absolute Performance }
\label{sec:5_1}

\begin{figure*}[t]
	\centering
	\subfloat[][FGSM adversarial attack ($\lambda = 0.3$)\label{fig:perturbations_fgsm3}]{\includegraphics[width=0.48\linewidth]{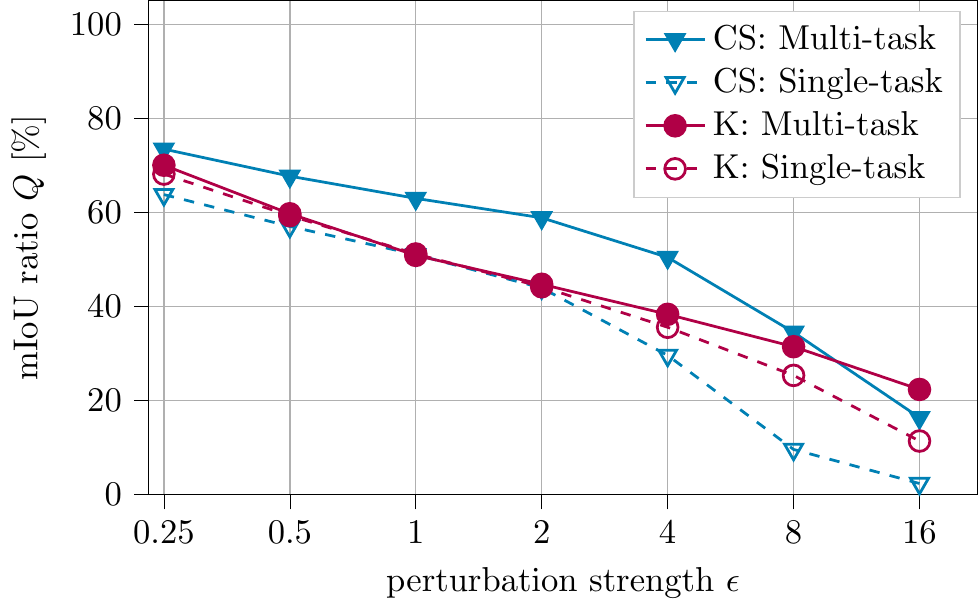}} \;\;%
	\subfloat[][FGSM adversarial attack ($\lambda = 0.6$)\label{fig:perturbations_fgsm6}]{\includegraphics[width=0.48\linewidth]{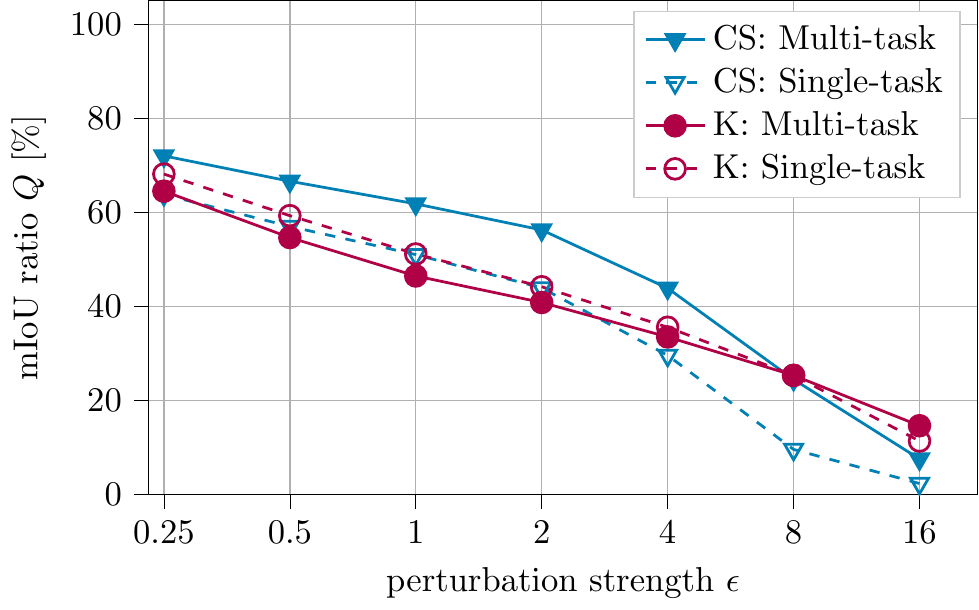}}
	\vspace{-0.1cm}
	\caption{Robustness of the two semantic segmentation models in terms of mIoU ratios $Q$ on the Cityscapes validation set (CS) and on the KITTI split's test set (K) for the \textbf{FGSM adversarial attack} for \textbf{different scale factors} $\lambda$.}
	\vspace{-0.1cm}
	\label{fig:ablation_perturbations}
\end{figure*}

Motivated by the question on how to properly weigh the decoder head's gradients against each other as described by (\ref{eq:gradient_scaling}), we trained models with different gradient scaling factors $\lambda$ and evaluated them on the Cityscapes validation set (CS) and the KITTI split's test set (K), cf.\ Table \ref{tab:comparison_dataset}. Note that here, as well as in Figures \ref{fig:scaling_factor}, \ref{fig:perturbations} and \ref{fig:ablation_perturbations}, we evaluate the same model on both datasets, which is why the Cityscapes performance represents the performance on the initial semantic segmentation dataset and the performance on the KITTI dataset expectedly shows somewhat weaker segmentation performance due to the domain shift. Figure \ref{fig:scaling_factor} shows that the performance on Cityscapes improves through multi-task training for all $\lambda \geq 0.1$, while on KITTI $\lambda = 0.1$ is best. This shows that our method can robustly improve the semantic segmentation performance on the original semantic segmentation dataset (Cityscapes), and for a good choice of $\lambda$ even on the dataset, only the depth is trained on (KITTI). Interestingly, a scaling factor of $\lambda = 1.0$ does not exhibit the (lower) performance of the baseline model trained only for the task of semantic segmentation (single-task), as although no gradients from the depth decoder are propagated into the encoder, the depth training images from another dataset still influence the statistics inside the batch normalization layers inside the encoder. This leads us to the conclusion that in our case the beneficial effect of multi-task learning is at least partly due to the easily accessible wide variety of data through a self-supervised task not relying on any labels.
\par 
Considering both datasets, we observe the best models using scaling factors of $0.1$ and $0.2$ with the values from Figure \ref{fig:scaling_factor} presented in Table \ref{tab:overall_performance}. We see that the model trained in a multi-task fashion improves by $5.4\%$ absolute over the baseline on Cityscapes and by $6.6\%$ absolute on KITTI. 
\begin{table}[t]
  \small
  \centering
  \setlength{\tabcolsep}{3pt}
  \begin{tabular}{|l|ccccccc|}
  \hline
  $\epsilon$ & $0.25$ & $0.5$ & $1$ & $2$ & $4$ & $8$ & $16$ \\
  \hline
  $Q$ (single-task) & $\textbf{100.0}$ & $\textbf{99.7}$ & $98.6$ & $93.7$ & $75.1$ & $44.7$ & $18.4$ \\
  $Q$ (multi-task) & $99.9$ & $\textbf{99.7}$ & $\textbf{99.4}$ & $\textbf{97.0}$ & $\textbf{87.5}$ & $\textbf{61.4}$ & $\textbf{28.0}$ \\
   \hline
  \end{tabular}
  \caption{Robustness of two semantic segmentation models under \textbf{Gaussian noise} input perturbations. For models trained with and without multi-task self-supervised depth learning, we present mIoU ratios $Q$ on the Cityscapes validation dataset for various perturbation strengths. mIoU ratio values in $\left[\%\right]$.}
  \label{tab:comparison_domain_shift}
\end{table}
\subsection{Robustness to Input Noise}
\label{sec:5_2}
\begin{figure*}[t]
	\centering
	\subfloat[][Gaussian noise\label{fig:perturbations_noise}]{\includegraphics[width=0.48\linewidth]{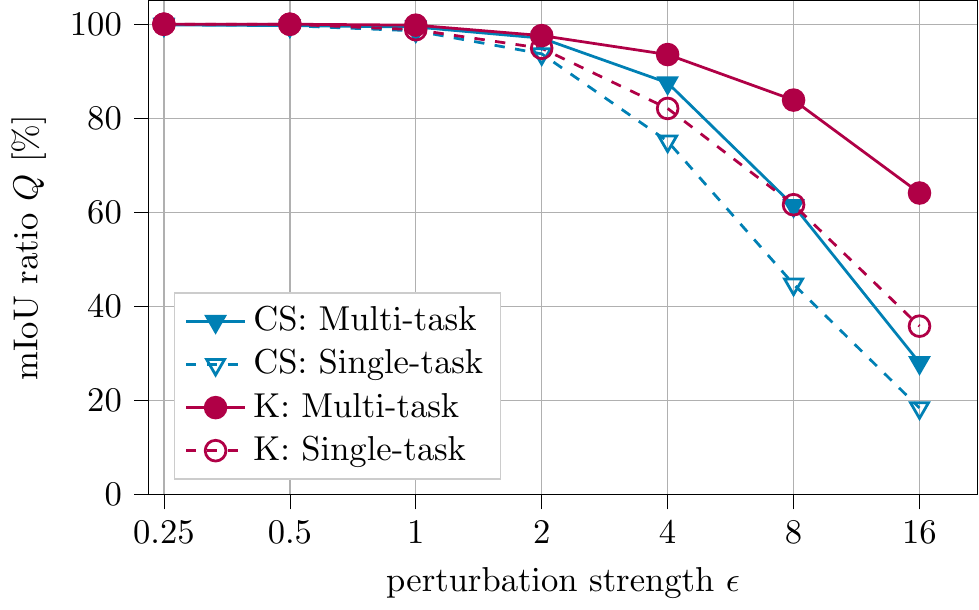}} \;\;%
	\subfloat[][Salt and pepper noise\label{fig:perturbations_brightness}]{\includegraphics[width=0.48\linewidth]{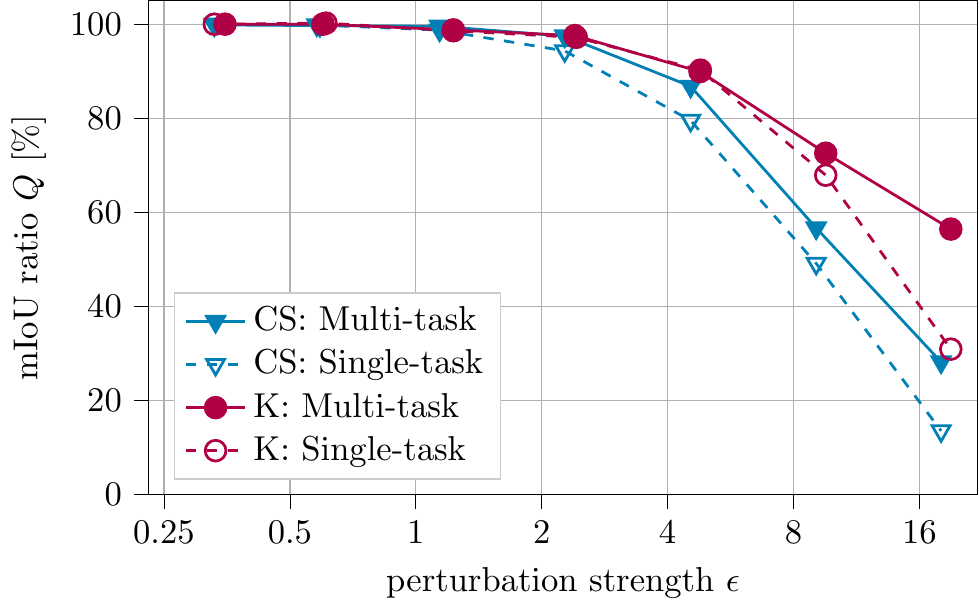}}\\[-0.1cm]
	\subfloat[][FGSM adversarial attack\label{fig:perturbations_attack1}]{\includegraphics[width=0.48\linewidth]{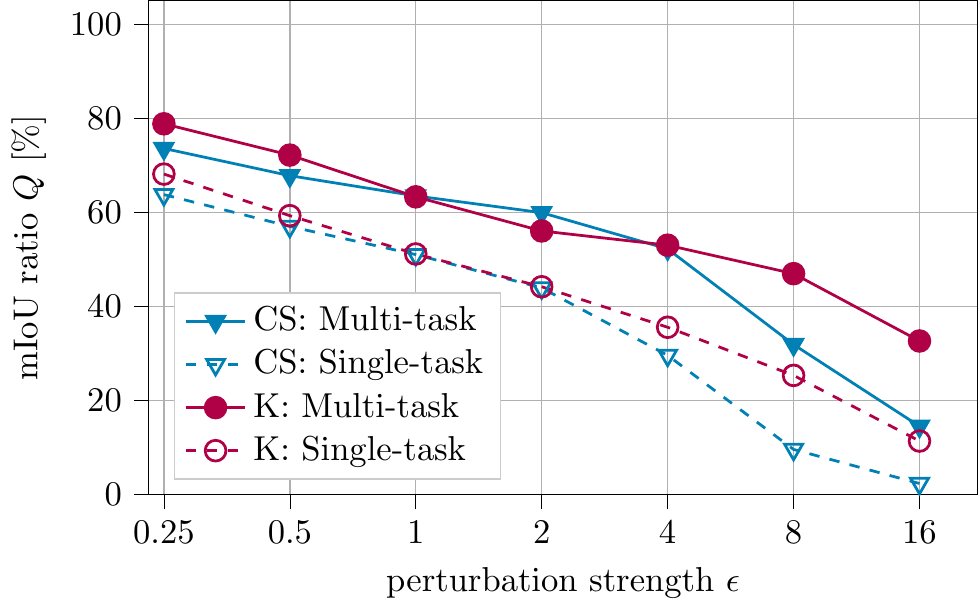}}\;\;%
	\subfloat[][PGD adversarial attack\label{fig:perturbations_attack2}]{\includegraphics[width=0.48\linewidth]{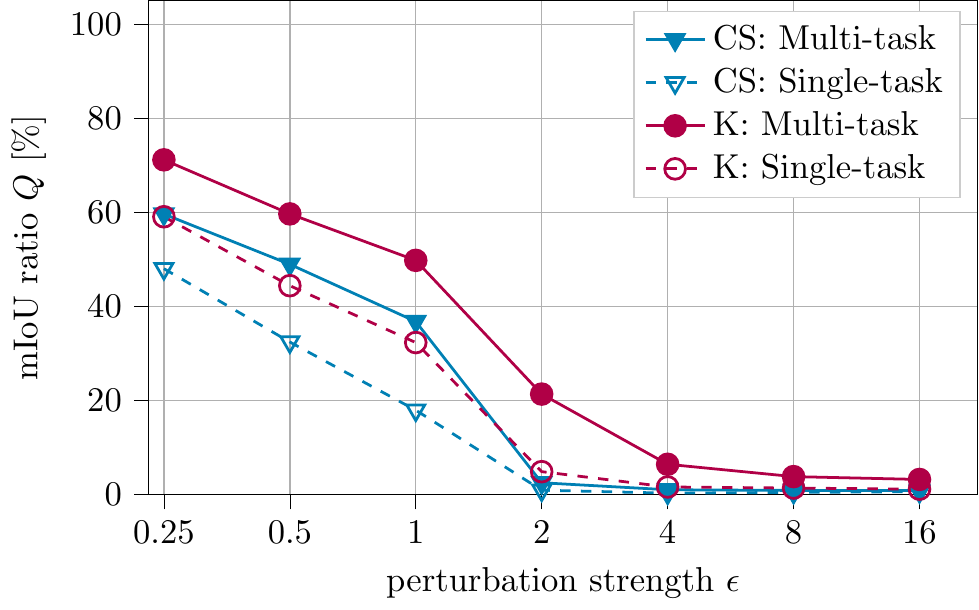}}
	\vspace{-0.1cm}
	\caption{Robustness of the two semantic segmentation models in terms of mIoU ratios $Q$ on the Cityscapes validation set (CS) and on the KITTI split's test set (K) for (a) \textbf{Gaussian noise}, (b) \textbf{salt and pepper noise}, as well as for (c) \textbf{FGSM} and (d) \textbf{PGD adversarial attacks}. The multi-task model was trained with $\lambda = 0.1$. All four subfigures can be mutually compared, since equal perturbation strengths $\epsilon$ in our experimental setting always means equal SNR of the input image $\boldsymbol{x}^{\mathrm{adv}}$ (power of $\boldsymbol{x}$ vs.\ the power of $\boldsymbol{r}$).}
	\label{fig:perturbations}
\end{figure*} 
\begin{table}[t]
  \small
  \centering
  \setlength{\tabcolsep}{4pt}
  \begin{tabular}{|l|ccccccc|}
  \hline
  $\epsilon$ & $0.25$ & $0.5$ & $1$ & $2$ & $4$ & $8$ & $16$ \\
  \hline
  $Q$ (single-task) & $63.8$ & $57.0$ & $51.0$ & $43.9$ & $29.6$ & $9.6$ & $2.4$ \\
  $Q$ (multi-task) & $\textbf{73.6}$ & $\textbf{67.8}$ & $\textbf{63.5}$ & $\textbf{59.9}$ & $\textbf{52.2}$ & $\textbf{31.9}$ & $\textbf{14.5}$ \\
   \hline
  \end{tabular}
  \caption{Robustness of the two semantic segmentation models to \textbf{FGSM adversarial attacks}. We present mIoU ratios $Q$ for models trained with and without multi-task self-supervised depth learning on the Cityscapes validation dataset for various perturbation strengths. mIoU ratio values in $\left[\%\right]$.}
  \label{tab:comparison_adversarial_attack}
\end{table}
While the improved performance for semantic segmentation is in accordance with previous works relying on stereo image pairs \cite{Chen2019a, Guizilini2020}, we also want to test the hypothesis that the additional self-supervised training can also improve robustness to input perturbations.\par
Starting to test this, we added Gaussian noise of different strengths to the input image during inference and evaluated, how the performance of a model trained only for the task of semantic segmentation (single-task) and a model that is trained in a multi-task fashion compare to each other. For this initial comparison we chose the best performing model with $\lambda=0.1$ as here already the positive effect of multi-task training with respect to absolute performance could be very effectively observed (cf.\ Fig.\ \ref{fig:scaling_factor}, Tab.\ \ref{tab:overall_performance}). A first overview of results is given in Table \ref{tab:comparison_domain_shift}, where one can see that Gaussian noise with small strength does not have much influence on the overall performance. At larger perturbation strengths ($\epsilon\geq 2$), the model trained in a multi-task fashion clearly retains a larger fraction of its performance.\par
Looking at Figures \ref{fig:perturbations_noise}, where each curve represents a row in Table \ref{tab:comparison_domain_shift}, and \ref{fig:perturbations_brightness}, we can observe that this improved robustness is obtained on both Cityscapes and KITTI for both Gaussian noise and salt and pepper noise. Particularly for strong input noises ($\epsilon\geq 2$), the robustness gain by multi-task learning is clearly observable. Concluding, we can state that \textit{the model trained in a multi-task fashion outperforms the single-task model in terms of robustness on both datasets and for both noise types}.
\subsection{Robustness to Adversarial Attacks}
\label{sec:5_3}
As so far the observations on input noise showed promising results, we also want to examine the case, where the input is deliberately perturbed by adversarial attacks. In Table \ref{tab:comparison_adversarial_attack} we show the same experiment as for the input Gaussian noise in Table \ref{tab:comparison_domain_shift}, but now we perturb the input images with FGSM adversarial patterns of different strengths. As a first observation, we note that for equal perturbation strengths $\epsilon$ (and due to our choice of (\ref{eq:epsilon}) also for equal SNRs) the mIoU ratio performance $Q$ is significantly lower as for random Gaussian noise. This is somehow expected as for adversarial examples the "noise" is optimized to fool the network. Furthermore, we can see that again \textit{the model trained in a multi-task fashion consistently outperforms the model trained solely for semantic segmentation}, indicating that the improved robustness seems to be a general characteristics of models trained in a multi-task fashion.\par
Again we wanted to challenge the robustness by more than one attack type and dataset, which is what we did in Figures \ref{fig:perturbations_attack1} and \ref{fig:perturbations_attack2}. \textit{For both FGSM and PGD adversarial attacks, we can see that the multi-task model clearly shows superior performance on both attack types and datasets}.

\subsection{Ablation Studies}
\label{sec:5_4}

After the successful experiments of Sections \ref{sec:5_2} and \ref{sec:5_3}, we also wanted to investigate the influence, the gradient scale factor $\lambda$ of (\ref{eq:gradient_scaling}) has on the robustness results. Therefore, we conducted the experiments on the FGSM attack also on the models trained with non-optimal scale factors $\lambda=0.3$ and $\lambda=0.6$, with results shown in Figures \ref{fig:perturbations_fgsm3} and \ref{fig:perturbations_fgsm6}. Interestingly, we observe that the gain in mIoU ratio $Q$ by multi-task training observed improved robustness on the Cityscapes dataset is quite independent of the scale factor, while this is not the case for KITTI.\footnote{We observed the same behavior for the other noise and attack types, but show the effect exemplarily for the FGSM attack.} We see that the segmentation performance as well as the robustness on the dataset the depth estimation is trained on can only be significantly improved for a good choice of the scale factor $\lambda$. However, remembering that the investigated models are initially trained on Cityscapes, and this is actually the dataset we are interested in \footnote{We do not claim domain transfer capabilities, although for a good choice of $\lambda$, even domain transferability is given to some extent by our multi-task learning approach as the good KITTI results in Fig.\ \ref{fig:scaling_factor} have shown.}, we find that \textit{one can very robustly improve the performance and robustness of a model on its original dataset (Cityscapes) through multi-task training with self-supervised monocular depth estimation}.

\section{Conclusions}
\label{sec:6}

In this work, we show the beneficial effects of multi-task training with self-supervised monocular depth estimation from videos on a semantic segmentation model's performance and robustness. Not only does the absolute performance improve, but also the robustness towards input perturbations improves, which indicates that one should utilize multi-task training across several tasks to obtain more robust models. We observed this behavior consistently on the dataset, the semantic segmentation is trained on in a supervised fashion, and for a good parameter choice also on the depth dataset. Employing a comparable scaling of perturbations, our evaluation bears the novelty to allow mutual robustness comparison even between noise and adversarial attack perturbations.\par
We see that the variety of data accessible by the self-supervised task is beneficial for the model to learn more robust features that are less sensitive to small input noise or adversarial attack perturbations. Our results indicate that our method offers an easy-to-implement mechanism to improve neural network robustness, while even improving the absolute performance at no additional computational overhead during inference.

\newpage
{\small
	\bibliographystyle{ieee_fullname}
	\bibliography{ifn_spaml_bibliography}

\begin{thebibliography}{10}\itemsep=-1pt

\bibitem{Arnab2018}
Anurag Arnab, Ondrej Miksik, and Philip H.~S. Torr.
\newblock {On the Robustness of Semantic Segmentation Models to Adversarial
  Attacks}.
\newblock In {\em Proc. of CVPR}, pages 888--897, Salt Lake City, UT, USA, June
  2018.

\bibitem{Assion2019}
Felix Assion, Peter Schlicht, Florens Greßner, Wiebke G\"{u}nther, Fabian
  H\"{u}ger, Nico~M. Schmidt, and Umair Rasheed.
\newblock {The Attack Generator: A Systematic Approach Towards Constructing
  Adversarial Attacks}.
\newblock In {\em Proc. of CVPR - Workshops}, pages 1--12, Long Beach, CA, USA,
  June 2019.

\bibitem{Athalye2018}
Anish Athalye, Nicholas Carlini, and David Wagner.
\newblock {Obfuscated Gradients Give a False Sense of Security: Circumventing
  Defenses to Adversarial Examples}.
\newblock In {\em Proc. of ICML}, pages 274--283, Stockholm, Sweden, July 2018.

\bibitem{Baer2019}
Andreas B\"{a}r, Fabian H\"{u}ger, Peter Schlicht, and Tim Fing\-scheidt.
\newblock {On the Robustness of Teacher-Student Frameworks for Semantic
  Segmentation}.
\newblock In {\em Proc. of CVPR - Workshops}, pages 1--9, Long Beach, CA, USA,
  June 2019.

\bibitem{Bilinski2018}
Piotr Bilinski and Victor Prisacariu.
\newblock {Dense Decoder Shortcut Connections for Single-Pass Semantic
  Segmentation}.
\newblock In {\em Proc. of CVPR}, pages 6596--6605, Salt Lake City, UT, USA,
  June 2018.

\bibitem{Bishop1995}
Chris~M. Bishop.
\newblock {Training With Noise is Equivalent to Tikhonov Regularization}.
\newblock {\em Neural Computation}, 7(1):108--116, Jan. 1995.

\bibitem{Bolte2019a}
Jan-Aike Bolte, Markus Kamp, Antonia Breuer, Silviu Homoceanu, Peter Schlicht,
  Fabian Huger, Daniel Lipinski, and Tim Fing\-scheidt.
\newblock {Unsupervised Domain Adaptation to Improve Image Segmentation Quality
  Both in the Source and Target Domain}.
\newblock In {\em Proc. of CVPR - Workshops}, pages 1--10, Long Beach, CA, USA,
  June 2019.

\bibitem{Bozorgtabar2019}
Behzad Bozorgtabar, Mohammad~Saeed Rad, Dwarikanath Mahapatra, and
  Jean-Philippe Thiran.
\newblock {SynDeMo: Synergistic Deep Feature Alignment for Joint Learning of
  Depth and Ego-Motion}.
\newblock In {\em Proc. of ICCV}, pages 4210--4219, Seoul, Korea, Oct. 2019.

\bibitem{RotaBulo2018}
Samuel~Rota Bul\`{o}, Lorenzo Porzi, and Peter Kontschieder.
\newblock {In-Place Activated BatchNorm for Memory-Optimized Training of DNNs}.
\newblock In {\em Proc. of CVPR}, pages 5639--5647, Salt Lake City, UT, USA,
  June 2018.

\bibitem{Carlini2017a}
Nicholas Carlini and David~A. Wagner.
\newblock {Towards Evaluating the Robustness of Neural Networks}.
\newblock In {\em Proc. of SP}, pages 39--57, San Jose, CA, USA, May 2017.

\bibitem{Casser2019}
Vincent Casser, Soeren Pirk, Reza Mahjourian, and Anelia Angelova.
\newblock {Depth Prediction Without the Sensors: Leveraging Structure for
  Unsupervised Learning from Monocular Videos}.
\newblock In {\em Proc. of AAAI}, pages 8001--8008, Honolulu, HI, USA, Jan.
  2019.

\bibitem{Chen2019e}
Hao-Yun Chen, Jhao-Hong Liang, Shih-Chieh Chang, Jia-Yu Pan, Yu-Ting Chen, Wei
  Wei, and Da-Cheng Juan.
\newblock {Improving Adversarial Robustness via Guided Complement Entropy}.
\newblock In {\em Proc. of ICCV}, pages 4881--4889, Seoul, Korea, Oct. 2019.

\bibitem{Chen2015}
Liang-Chieh Chen, George Papandreou, Iasonas Kokkinos, Kevin Murphy, and
  Alan~L. Yuille.
\newblock {Semantic Image Segmentation With Deep Convolutional Nets and Fully
  Connected CRFs}.
\newblock In {\em Proc. of ICLR}, pages 1--14, San Diego, CA, USA, May 2015.

\bibitem{Chen2018}
Liang-Chieh Chen, George Papandreou, Iasonas Kokkinos, Kevin Murphy, and
  Alan~L. Yuille.
\newblock {DeepLab: Semantic Image Segmentation With Deep Convolutional Nets,
  Atrous Convolution, and Fully Connected CRFs}.
\newblock {\em IEEE Transactions on Pattern Analysis and Machine Intelligence},
  40(4):834--848, Apr. 2018.

\bibitem{Chen2018a}
Liang-Chieh Chen, Yukun Zhu, George Papandreou, Florian Schroff, and Hartwig
  Adam.
\newblock {Encoder-Decoder With Atrous Separable Convolution for Semantic Image
  Segmentation}.
\newblock In {\em Proc. of ECCV}, pages 801--818, Munich, Germany, Sept. 2018.

\bibitem{Chen2019a}
Po-Yi Chen, Alexander~H. Liu, Yen-Cheng Liu, and Yu-Chiang~F. Wang.
\newblock {Towards Scene Understanding: Unsupervised Monocular Depth Estimation
  With Semantic-Aware Representation}.
\newblock In {\em Proc. of CVPR}, pages 2624--2632, Long Beach, CA, USA, June
  2019.

\bibitem{Chen2019b}
Yuhua Chen, Cordelia Schmid, and Cristian Sminchisescu.
\newblock {Self-Supervised Learning With Geometric Constraints in Monocular
  Video Connecting Flow, Depth, and Camera}.
\newblock In {\em Proc. of ICCV}, pages 7063--7072, Seoul, Korea, Oct. 2019.

\bibitem{Chollet2017}
Fran\c{c}ois Chollet.
\newblock {Xception: Deep Learning With Depthwise Separable Convolutions}.
\newblock In {\em Proc. of CVPR}, pages 1063--6919, Honolulu, HI, USA, July
  2017.

\bibitem{Cordts2016}
Marius Cordts, Mohamed Omran, Sebastian Ramos, Timo Rehfeld, Markus Enzweiler,
  Rodrigo Benenson, Uwe Franke, Stefan Roth, and Bernt Schiele.
\newblock {The Cityscapes Dataset for Semantic Urban Scene Understanding}.
\newblock In {\em Proc. of CVPR}, pages 3213--3223, Las Vegas, NV, USA, June
  2016.

\bibitem{Dong2018a}
Yinpeng Dong, Fangzhou Liao, Tianyu Pang, Hang Su, Jun Zhu, Xiaolin Hu, and
  Jianguo Li.
\newblock {Boosting Adversarial Attacks With Momentum}.
\newblock In {\em Proc. of CVPR}, pages 9185--9193, Salt Lake City, UT, USA,
  June 2018.

\bibitem{Eigen2014}
David Eigen, Christian Puhrsch, and Rob Fergus.
\newblock {Depth Map Prediction from a Single Image Using a Multi-Scale Deep
  Network}.
\newblock In {\em Proc. of NIPS}, pages 2366--2374, Montr\' {e}al, QC, Canada,
  Dec. 2014.

\bibitem{Everingham2015}
Mark Everingham, Luc Van~Gool, Christopher K.~I. Williams, John Winn, and
  Andrew Zisserman.
\newblock {The Pascal Visual Object Classes Challenge: A Retrospective}.
\newblock {\em International Journal of Computer Vision (IJCV)},
  111(1):98--136, Jan. 2015.

\bibitem{Fu2018}
Huan Fu, Mingming Gong, Chaohui Wang, Kayhan Batmanghelich, and Dacheng Tao.
\newblock {Deep Ordinal Regression Network for Monocular Depth Estimation}.
\newblock In {\em Proc. of CVPR}, pages 2002--2011, Salt Lake City, UT, USA,
  June 2018.

\bibitem{Ganin2015}
Yaroslav Ganin and Victor Lempitsky.
\newblock {Unsupervised Domain Adaptation by Backpropagation}.
\newblock In {\em Proc. of ICML}, pages 1180--1189, Lille, France, July 2015.

\bibitem{Garg2016}
Ravi Garg, Vijay~Kumar BG, Gustavo Carneiro, and Ian Reid.
\newblock {Unsupervised CNN for Single View Depth Estimation: Geometry to the
  Rescue}.
\newblock In {\em Proc. of ECCV}, pages 740--756, Amsterdam, The Netherlands,
  Oct. 2016.

\bibitem{Geiger2013}
Andreas Geiger, Philip Lenz, Christoph Stiller, and Raquel Urtasun.
\newblock {Vision Meets Robotics: The KITTI Dataset}.
\newblock {\em International Journal of Robotics Research (IJRR)},
  32(11):1231--1237, Aug. 2013.

\bibitem{Godard2017}
Cl{\'e}ment Godard, Oisin Mac~Aodha, and Gabriel~J. Brostow.
\newblock {Unsupervised Monocular Depth Estimation With Left-Right
  Consistency}.
\newblock In {\em Proc. of CVPR}, pages 270--279, Honolulu, HI, USA, July 2017.

\bibitem{Godard2019}
Cl{\'e}ment Godard, Oisin Mac~Aodha, Michael Firman, and Gabriel~J. Brostow.
\newblock {Digging Into Self-Supervised Monocular Depth Estimation}.
\newblock In {\em Proc. of ICCV}, pages 3828--3838, Seoul, Korea, Oct. 2019.

\bibitem{Goodfellow2015}
Ian Goodfellow, Jonathon Shlens, and Christian Szegedy.
\newblock {Explaining and Harnessing Adversarial Examples}.
\newblock In {\em Proc. of ICLR}, pages 1--10, San Diego, CA, USA, May 2015.

\bibitem{Gordon2019}
Ariel Gordon, Hanhan Li, Rico Jonschkowski, and Anelia Angelova.
\newblock {Depth from Videos in the Wild: Unsupervised Monocular Depth Learning
  from Unknown Cameras}.
\newblock In {\em Proc. of ICCV}, pages 8977--8986, Seoul, Korea, Oct. 2019.

\bibitem{Guizilini2019}
Vitor Guizilini, Rares Ambrus, Sudeep Pillai, and Adrien Gaidon.
\newblock {PackNet-SfM: 3D Packing for Self-Supervised Monocular Depth
  Estimation}.
\newblock {\em arXiv}, (1905.02693), May 2019.

\bibitem{Guizilini2020}
Vitor Guizilini, Rui Hou, Jie Li, Rares Ambrus, and Adrien Gaidon.
\newblock {Semantically-Guided Representation Learning for Self-Supervised
  Monocular Depth}.
\newblock In {\em Proc. of ICLR}, pages 1--14, Addis Ababa, Ethiopia, Apr.
  2020.

\bibitem{Guizilini2019a}
Vitor Guizilini, Jie Li, Rares Ambrus, Sudeep Pillai, and Adrien Gaidon.
\newblock {Robust Semi-Supervised Monocular Depth Estimation With Reprojected
  Distances}.
\newblock In {\em Proc. of CoRL}, pages 1--14, Osaka, Japan, Oct. 2019.

\bibitem{Guo2018}
Chuan Guo, Mayank Rana, Moustapha Ciss\'{e}, and Laurens van~der Maaten.
\newblock {Countering Adversarial Images using Input Transformations}.
\newblock In {\em Proc. of ICLR}, pages 1--12, Vancouver, BC, Canada, Apr.
  2018.

\bibitem{He2016}
Kaiming He, Xiangyu Zhang, Shaoqing Ren, and Jian Sun.
\newblock {Deep Residual Learning for Image Recognition}.
\newblock In {\em Proc. of CVPR}, pages 770--778, Las Vegas, NV, USA, June
  2016.

\bibitem{He2019c}
Zhezhi He, Adnan~S. Rakin, and Deliang Fan.
\newblock {Parametric Noise Injection: Trainable Randomness to Improve Deep
  Neural Network Robustness Against Adversarial Attack}.
\newblock In {\em Proc. of CVPR}, pages 588--597, Long Beach, CA, USA, June
  2019.

\bibitem{Hendrycks2019a}
Dan Hendrycks, Mantas Mazeika, Saurav Kadavath, and Dawn Song.
\newblock {Using Self-Supervised Learning Can Improve Model Robustness and
  Uncertainty}.
\newblock In {\em Proc. of NeurIPS}, pages 15637--15648, Vancouver, BC, Canada,
  Dec. 2019.

\bibitem{Holstroem1992}
Lasse Holstr\"{o}m and Petri Koistinen.
\newblock {Using Additive Noise in Backpropagation-Training}.
\newblock {\em IEEE Transactions on Neural Networks}, 3(1):24--38, Jan. 1992.

\bibitem{Kendall2018}
Alex Kendall, Yarin Gal, and Roberto Cipolla.
\newblock {Multi-Task Learning Using Uncertainty to Weigh Losses for Scene
  Geometry and Semantics}.
\newblock In {\em Proc. of CVPR}, pages 7482--7491, Salt Lake City, UT, USA,
  June 2018.

\bibitem{Kingma2015}
Diederik~P. Kingma and Jimmy Ba.
\newblock {Adam: A Method for Stochastic Optimization}.
\newblock In {\em Proc. of ICLR}, pages 1--15, San Diego, CA, USA, May 2015.

\bibitem{Kurakin2017}
Alexey Kurakin, Ian Goodfellow, and Samy Bengio.
\newblock {Adversarial Examples in the Physical World}.
\newblock In {\em Proc. of ICLR - Workshops}, pages 1--14, Toulon, France, Apr.
  2017.

\bibitem{Li2019a}
Hanchao Li, Pengfei Xiong, Haoqiang Fan, and Jian Sun.
\newblock {DFANet: Deep Feature Aggregation for Real-Time Semantic
  Segmentation}.
\newblock In {\em Proc. of CVPR}, pages 9522--9531, Long Beach, CA, USA, June
  2019.

\bibitem{Liu2019e}
Hong Liu, Rongrong Ji, Jie Li, Baochang Zhang, Yue Gao, Yongjian Wu, and Feiyue
  Huang.
\newblock {Universal Adversarial Perturbation via Prior Driven Uncertainty
  Approximation}.
\newblock In {\em Proc. of ICCV}, pages 2941--2949, Seoul, Korea, Oct. 2019.

\bibitem{Liu2019c}
Zihao Liu, Qi Liu, Tao Liu, Nuo Xu, Xue Lin, Yanzhi Wang, and Wujie Wen.
\newblock {Feature Distillation: DNN-Oriented JPEG Compression Against
  Adversarial Examples}.
\newblock In {\em Proc. of CVPR}, pages 860--868, Long Beach, CA, USA, June
  2019.

\bibitem{Long2015}
Jonathan Long, Evan Shelhamer, and Trevor Darrell.
\newblock {Fully Convolutional Networks for Semantic Segmentation}.
\newblock In {\em Proc. of CVPR}, pages 3431--3440, Boston, MA, USA, June 2015.

\bibitem{Madry2018}
Aleksander Madry, Aleksandar Makelov, Ludwig Schmidt, Dimitris Tsipras, and
  Adrian Vladu.
\newblock {Towards Deep Learning Models Resistant to Adversarial Attacks}.
\newblock In {\em Proc. of ICLR}, pages 1--28, Vancouver, BC, Canada, Apr.
  2018.

\bibitem{Mahjourian2018}
Reza Mahjourian, Martin Wicke, and Anelia Angelova.
\newblock {Unsupervised Learning of Depth and Ego-Motion from Monocular Video
  Using 3D Geometric Constraints}.
\newblock In {\em Proc. of CVPR}, pages 5667--5675, Salt Lake City, UT, USA,
  June 2018.

\bibitem{Mehta2018}
Sachin Mehta, Mohammad Rastegari, Anat Caspi, Linda Shapiro, and Hannaneh
  Hajishirzi.
\newblock {ESPNet: Efficient Spatial Pyramid of Dilated Convolutions for
  Semantic Segmentation}.
\newblock In {\em Proc. of ECCV}, pages 552--568, Munich, Germany, Sept. 2018.

\bibitem{Meng2019a}
Yue Meng, Yongxi Lu, Aman Raj, Samuel Sunarjo, Rui Guo, Tara Javidi, Gaurav
  Bansal, and Dinesh Bharadia.
\newblock {SIGNet: Semantic Instance Aided Unsupervised 3D Geometry
  Perception}.
\newblock In {\em Proc. of CVPR}, pages 9810--9820, Long Beach, CA, USA, June
  2019.

\bibitem{Menze2015}
Moritz Menze and Andreas Geiger.
\newblock {Object Scene Flow for Autonomous Vehicles}.
\newblock In {\em Proc. of CVPR}, pages 3061--3070, Boston, MA, USA, June 2015.

\bibitem{Metzen2017}
Jan~H. Metzen, Mummadi~C. Kumar, Thomas Brox, and Volker Fischer.
\newblock {Universal Adversarial Perturbations Against Semantic Image
  Segmentation}.
\newblock In {\em Proc. of ICCV}, pages 2774--2783, Venice, Italy, Oct. 2017.

\bibitem{Moosavi-Dezfooli2017}
Seyed-Mohsen Moosavi-Dezfooli, Alhussein Fawzi, Omar Fawzi, and Pascal
  Frossard.
\newblock {Universal Adversarial Perturbations}.
\newblock In {\em Proc. of CVPR}, pages 1765--1773, Honolulu, HI, USA, July
  2017.

\bibitem{Mopuri2018}
Konda~R. Mopuri, Aditya Ganeshan, and Venkatesh~B. Radhakrishnan.
\newblock {Generalizable Data-Free Objective for Crafting Universal Adversarial
  Perturbations}.
\newblock {\em IEEE Transactions on Pattern Analysis and Machine Intelligence
  (TPAMI)}, 41(10):2452--2465, Oct. 2019.

\bibitem{Mopuri2017}
Konda~Reddy Mopuri, Utsav Garg, and R.~Venkatesh Babu.
\newblock {Fast Feature Fool: A Data Independent Approach to Universal
  Adversarial Perturbations}.
\newblock In {\em Proc. of BMVC}, pages 1--12, London, UK, Sept. 2017.

\bibitem{Novosel2019}
Jelena Novosel, Prashanth Viswanath, and Bruno Arsenali.
\newblock {Boosting Semantic Segmentation With Multi-Task Self-Supervised
  Learning for Autonomous Driving Applications}.
\newblock In {\em Proc. of NeurIPS - Workshops}, pages 1--11, Vancouver, BC,
  Canada, Dec. 2019.

\bibitem{Orsic2019}
Marin Or\u{s}i\'{c}, Ivan Kre\u{s}o, Petra Bevandi\'{c}, and Sini\u{s}a
  \u{S}egvi\'{c}.
\newblock {In Defense of Pre-Trained ImageNet Architectures for Real-Time
  Semantic Segmentation of Road-Driving Images}.
\newblock In {\em Proc. of CVPR}, pages 12607--12616, Long Beach, CA, USA, June
  2019.

\bibitem{Paszke2016}
Adam Paszke, Abhishek Chaurasia, Sangpil Kim, and Eugenio Culurciello.
\newblock {ENet: A Deep Neural Network Architecture for Real-Time Semantic
  Segmentation}.
\newblock {\em arXiv}, June 2016.
\newblock (arXiv:1606.02147).

\bibitem{Paszke2019}
Adam Paszke, Sam Gross, Francisco Massa, Adam Lerer, James Bradbury, Gregory
  Chanan, Trevor Killeen, Zeming Lin, Natalia Gimelshein, Luca Antiga, Alban
  Desmaison, Andreas Kopf, Edward Yang, Zachary DeVito, Martin Raison, Alykhan
  Tejani, Sasank Chilamkurthy, Benoit Steiner, Lu Fang, Junjie Bai, and Soumith
  Chintala.
\newblock {PyTorch: An Imperative Style, High-Performance Deep Learning
  Library}.
\newblock In {\em Proc. of NeurIPS}, pages 8024--8035, Vancouver, BC, Canada,
  Dec. 2019.

\bibitem{Pillai2019}
Sudeep Pillai, Rare{\c{s}} Ambru{\c{s}}, and Adrien Gaidon.
\newblock {SuperDepth: Self-Supervised, Super-Resolved Monocular Depth
  Estimation}.
\newblock In {\em Proc. of ICRA}, pages 9250--9256, Montr{\' e}al, QC, Canada,
  May 2019.

\bibitem{Romera2018}
Eduardo Romera, Jos\'{e}~M. \'{A}lvarez, Luis~M. Bergasa, and Roberto Arroyo.
\newblock {ERFNet: Efficient Residual Factorized Conv\-Net for Real-Time
  Semantic Segmentation}.
\newblock {\em IEEE Transactions on Intelligent Transportation Systems},
  19(1):263--272, Jan. 2018.

\bibitem{Russakovsky2015}
Olga Russakovsky, Jia Deng, Hao Su, Jonathan Krause, Sanjeev Satheesh, Sean Ma,
  Zhiheng Huang, Andrej Karpathy, Aditya Khosla, Michael Bernstein,
  Alexander~C. Berg, and Li Fei-Fei.
\newblock {ImageNet Large Scale Visual Recognition Challenge}.
\newblock {\em International Journal of Computer Vision (IJCV)},
  115(3):211--252, Dec. 2015.

\bibitem{Sandler2018}
Mark Sandler, Andrew~G. Howard, Menglong Zhu, Andrey Zhmoginov, and Liang-Chieh
  Chen.
\newblock {MobileNetV2: Inverted Residuals and Linear Bottlenecks}.
\newblock In {\em Proc. of CVPR}, pages 4510--4520, Salt Lake City, UT, USA,
  June 2018.

\bibitem{Szegedy2014}
Christian Szegedy, Wojciech Zaremba, Ilya Sutskever, Joan Bruna, Dumitru Erhan,
  Ian Goodfellow, and Rob Fergus.
\newblock {Intriguing Properties of Neural Networks}.
\newblock In {\em Proc. of ICLR}, pages 1--10, Montr\' {e}al, QC, Canada, Dec.
  2014.

\bibitem{Szeliski2010}
Richard Szeliski.
\newblock {\em {Computer Vision: Algorithms and Applications}}.
\newblock {Springer Science \& Business Media}, 2010.

\bibitem{Vemulapalli2016}
Raviteja Vemulapalli, Oncel Tuzel, Ming-Yu Liu, and Rama Chellappa.
\newblock {Gaussian Conditional Random Field Network for Semantic
  Segmentation}.
\newblock In {\em Proc. of CVPR}, pages 3224--3233, Las Vegas, NV, USA, June
  2016.

\bibitem{Vu2019a}
Tuan-Hung Vu, Himalaya Jain, Maxime Bucher, Matthieu Cord, and Patrick
  P\'{e}rez.
\newblock {DADA: Depth-Aware Domain Adaptation in Semantic Segmentation}.
\newblock In {\em Proc. of ICCV}, pages 7364--7373, Seoul, Korea, Oct. 2019.

\bibitem{wang2004image}
Zhou Wang, Alan~Conrad Bovik, Hamid~Rahim Sheikh, and Eero~P. Simoncelli.
\newblock {Image Quality Assessment: From Error Visibility to Structural
  Similarity}.
\newblock {\em IEEE Trans. on Image Processing}, 13(4):600--612, Apr. 2004.

\bibitem{Xu2018c}
Dan Xu, Wanli Ouyang, Xiaogang Wang, and Nicu Sebe.
\newblock {PAD-Net: Multi-Tasks Guided Prediction-and-Distillation Network for
  Simultaneous Depth Estimation and Scene Parsing}.
\newblock In {\em Proc. of CVPR}, pages 675--684, Salt Lake City, UT, USA, June
  2018.

\bibitem{Yang2018b}
Guorun Yang, Hengshuang Zhao, Jianping Shi, Zhidong Deng, and Jiaya Jia.
\newblock {SegStereo: Exploiting Semantic Information for Disparity
  Estimation}.
\newblock In {\em Proc. of ECCV}, pages 636--651, Munich, Germany, Sept. 2018.

\bibitem{Yin2018}
Zhichao Yin and Jianping Shi.
\newblock {GeoNet: Unsupervised Learning of Dense Depth, Optical Flow and
  Camera Pose}.
\newblock In {\em Proc. of CVPR}, pages 1983--1992, Salt Lake City, UT, USA,
  June 2018.

\bibitem{Yu2016}
Fisher Yu and Vladlen Koltun.
\newblock {Multi-Scale Context Aggregation by Dilated Convolutions}.
\newblock In {\em Proc. of ICLR}, pages 1--13, San Juan, Puerto Rico, May 2016.

\bibitem{Zhang2019a}
Zhenyu Zhang, Zhen Cui, Chunyan Xu, Yan Yan, Nicu Sebe, and Jian Yang.
\newblock {Pattern-Affinitive Propagation Across Depth, Surface Normal and
  Semantic Segmentation}.
\newblock In {\em Proc. of CVPR}, pages 4106--4115, Long Beach, CA, USA, June
  2019.

\bibitem{Zhao2016a}
Hengshuang Zhao, Jianping Shi, Xiaojuan Qi, Xiaogang Wang, and Jiaya Jia.
\newblock {Pyramid Scene Parsing Network}.
\newblock In {\em Proc. of CVPR}, pages 2881--2890, Honulu, HI, USA, July 2017.

\bibitem{Zhou2017a}
Tinghui Zhou, Matthew Brown, Noah Snavely, and David~G. Lowe.
\newblock {Unsupervised Learning of Depth and Ego-Motion from Video}.
\newblock In {\em Proc. of CVPR}, pages 1851--1860, Honolulu, HI, USA, July
  2017.

\bibitem{Zhu2019}
Yi Zhu, Karan Sapra, Fitsum~A. Reda, Kevin~J. Shih, Shawn Newsam, Andrew Tao,
  and Bryan Catanzaro.
\newblock {Improving Semantic Segmentation via Video Propagation and Label
  Relaxation}.
\newblock In {\em Proc. of CVPR}, pages 8856--8865, Long Beach, CA, USA, June
  2019.

\bibitem{Zhuang2018}
Yueqing Zhuang, Fan Yang, Li Tao, Cong Ma, Ziwei Zhang, Yuan Li, Huizhu Jia,
  Xiaodong Xie, and Wen Gao.
\newblock {Dense Relation Network: Learning Consistent and Context-Aware
  Representation for Semantic Image Segmentation}.
\newblock In {\em Proc. of ICIP}, pages 3698--3702, Athens, Greece, Oct. 2018.

\bibitem{Zou2018a}
Yuliang Zou, Zelun Luo, and Jia-Bin Huang.
\newblock {DF-Net: Unsupervised Joint Learning of Depth and Flow Using
  Cross-Task Consistency}.
\newblock In {\em Proc. of ECCV}, pages 36--53, Munich, Germany, Sept. 2018.

\bibitem{Zou2018}
Yang Zou, Zhiding Yu, B.~V.~K. Vijaya~Kumar, and Jinsong Wang.
\newblock {Unsupervised Domain Adaptation for Semantic Segmentation via
  Class-Balanced Self-Training}.
\newblock In {\em Proc. of ECCV}, pages 289--305, Munich, Germany, Sept. 2018.

\end{thebibliography}
}

\end{document}